\documentclass[lettersize,journal]{IEEEtran}
\usepackage{amsmath,amsfonts}
\usepackage{amssymb}
\usepackage{algorithmic}
\usepackage{algorithm}
\usepackage{array}
\usepackage[caption=false,font=normalsize,labelfont=sf,textfont=sf]{subfig}
\usepackage{textcomp}
\usepackage{stfloats}
\usepackage{url}
\usepackage{verbatim}
\usepackage{graphicx}
\usepackage{cite}
\usepackage{multirow}
\usepackage{pdfpages}
\usepackage{tabularx}
\usepackage{hyperref}
\usepackage{float}
\usepackage{stfloats}

\begin{document}

\title{EFLNet: Enhancing Feature Learning Network for Infrared Small Target Detection}

\author{Bo Yang, Xinyu Zhang, Jian Zhang, Jun Luo, Mingliang Zhou,~\IEEEmembership{Member,~IEEE,} Yangjun Pi{}
\thanks{This work was supported in part by the Fundamental Research Funds for the Central Universities under Grant 2023CDJXY-021. (\emph{Corresponding author: Yangjun Pi)}}
\thanks{Bo Yang,  Xinyu Zhang, Jian Zhang, Jun Luo and Yangjun Pi are with the  State  Key  Laboratory  of Mechanical Transmission for Advanced Equipment, College of Mechanical and Vehicle Engineering, Chongqing University, Chongqing, 400044, China  (e-mail:cqpp@cqu.edu.cn).}
\thanks{Mingliang Zhou is with the School of Computer Science, Chongqing University, Chongqing, 400044, China.}}

\markboth{Under Submission}%
{Shell \MakeLowercase{\textit{et al.}}: A Sample Article Using IEEEtran.cls for IEEE Journals}


\maketitle

\begin{abstract}
Single-frame infrared small target detection is considered to be a challenging task, due to the extreme imbalance between target and background, bounding box regression is extremely sensitive to infrared small target, and target information is easy to lose in the high-level semantic layer. In this paper, we propose an enhancing feature learning network (EFLNet) to address these problems. First, we notice that there is an extremely imbalance between the target and the background in the infrared image, which makes the model pay more attention to the background features rather than target features. To address this problem, we propose a new adaptive threshold focal loss (ATFL) function that decouples the target and the background, and utilizes the adaptive mechanism to adjust the loss weight to force the model to allocate more attention to target features. Second, we introduce the normalized Gaussian Wasserstein distance (NWD) to alleviate the difficulty of convergence caused by the extreme sensitivity of the bounding box regression to infrared small target. Finally, we incorporate a dynamic head mechanism into the network to enable adaptive learning of the relative importance of each semantic layer. Experimental results demonstrate our method can achieve better performance in the detection performance of infrared small target compared to state-of-the-art deep-learning based methods. The source codes and bounding box annotated datasets are
available at \textcolor{red}{https://github.com/YangBo0411/infrared-small-target}.
\end{abstract}

\begin{IEEEkeywords}
Infrared small target detection, deep learning, adaptive threshold focal loss, dynamic head.
\end{IEEEkeywords}

\section{Introduction}
\IEEEPARstart{I}{nfrared} small target detection serves a crucial role in various applications, including ground monitoring\cite{35}, early warning systems\cite{1}, precision guidance\cite{2}, and others. In comparison to conventional object detection tasks, infrared small target detection exhibits distinct characteristics. First, due to the target's size or distance, the proportion of the target within the infrared image is exceedingly small, often comprising just a few pixels or a single pixel in extreme cases. Second, the objects in infrared small target detection tasks are typically sparsely distributed, usually containing only one or a few instances, each of which occupies a minuscule portion of the entire image. As a result, a significant imbalance arises between the target area and the background area. Moreover, the background of infrared small target is intricate, containing substantial amounts of noise and exhibiting a low signal-to-clutter ratio (SCR). Consequently, the target becomes prone to being overshadowed by the background. These distinctive features render infrared small target detection exceptionally challenging.

Various model-based methods have been proposed for infrared small target detection, including filter-based methods\cite{3,4}, local contrast-based methods\cite{5,6}, and low-rank-based methods\cite{7,8}. The filter-based methods segment the target by estimating the background and enhancing the target. However, their suitability are limited to even backgrounds, and they lack robustness when faced with complex backgrounds. The local contrast-based methods identify the target by calculating the intensity difference between the target and its surrounding neighborhood. Nevertheless, they struggle to effectively detect dim targets. The low-rank decomposition methods distinguish the structural features of the target and background based on the sparsity of the target and the low-rank characteristics of the background. Nonetheless, they exhibit a high false alarm rate when confronted with images featuring complex background and variations in target shape. In practical scenarios, infrared images often exhibit complex background, dim targets, and a low SCR, which poses a possibility of failure for these methods.

In recent years, deep learning has witnessed remarkable advancements, leading to significant breakthroughs in numerous domains. In contrast to traditional methods for infrared small target detection, deep learning leverage a data-driven end-to-end learning framework, enabling adaptive feature learning of infrared small target without the need for manual feature making. Since the work of miss detection vs. false alarm (MDvsFA)\cite{9} and asymmetric contextual modulation networks (ACMNet)\cite{10}, some deep-learning based methods have been proposed. Despite the notable achievements of existing deep-learning based methods in infrared small target detection, the majority of current research treats it as a segmentation task\cite{36,37,38,40}. The segmentation tasks offer pixel-level detailed information, which is advantageous in scenarios that demand precise differentiation. However, segmentation tasks necessitate the processing of pixel-level details, requiring substantial computational resources. Consequently, the training and inference times tend to be prolonged. In addition, semantic segmentation is only an intermediate representation, which is used as input to track and locate infrared small target, and segmentation integrity is only an approximation of detection accuracy, and the specific detection performance cannot be evaluated. Therefore, there have been works to model infrared small target detection as an object detection problem\cite{23,25,34,39}.

However, the detection performance of infrared small target remains insufficient compared to the detection of normal targets. This inadequacy can be attributed to three key factors. First, the imbalance between the target and background in the image will cause the detector to learn more background information and tend to mistakenly recognize the target as the background, while not paying enough attention to the target information. Second, infrared small target are highly sensitive to the intersection over union (IOU) metric, rendering precise bounding box regression challenging as even slight changes in the bounding box can significantly impact the IOU calculation. Third, the information of infrared small target is easily lost during the downsampling process, and shallow features containing more target information are not taken seriously.

To address above problem, this paper proposes a detection-based method called enhancing feature learning network (EFLNet), which can improve the detection performance of infrared small target. First, we design the adaptive threshold focal loss function (ATFL) to alleviate the imbalance problem between the target and the background in the infrared image. Furthermore, to achieve more accurate bounding box regression for infrared small target, a two-dimensional Gaussian distribution is used to remodel the bounding box, and the normalized Gaussian Wasserstein distance (NWD) is employed to address the problem of infrared small target being highly sensitive to IoU. Finally, we incorporate a dynamic head into the detection network. The relative importance of each semantic layer is learned through the self-attention mechanism, which improves the detection performance of infrared small target. Furthermore, most of the existing infrared small target datasets solely offer mask annotation versions,  limiting the scope of infrared small target detection to a segmentation task. We provide the corresponding bounding box annotation versions for the current infrared small target public datasets,  which makes it possible to make infrared small target detection as a detection-based task.

Our contributions can be summarized as follows:

• We propose an EFLNet to improve the detection performance of infrared small target. The feature learning ability of the network for infrared small target can be well enhanced by more suitable loss function and network structure.

• We designed an ATFL for infrared small target, which can decouple the target from the background and dynamically adjust the loss weight, allowing the model to assign greater attention to hard-to-detect targets.

• We provide a bounding box annotation version of the current infrared small target public dataset, which makes up for the lack of bounding box annotation version in the current dataset and facilitates the detection task.

The remainder of this paper is organized as follows: Section \ref{section2} provides related work of the existing research on infrared small target detection. Section \ref{section3} introduces the proposed network architecture. The experimental results and analysis are presented in Section \ref{section4}. Finally, Section \ref{section5} concludes the entire article.
\section{Related work}\label{section2}
\subsection{Model-based method}
Extensive research was conducted by researchers to address the problem of infrared small target detection. Filter-based methods, such as MaxMedian\cite{3}, Tophat\cite{4}, two-dimensional adaptive least-mean-square (TDLMS)\cite{11}, and two-dimensional variational mode decomposition (TDVMD)\cite{12}, demonstrated good performance on smooth or low-frequency backgrounds but exhibited limitations when dealed with complex backgrounds. Local-contrast based methods like weighted strengthened local contrast measure (WSLCM)\cite{1}, tri-layer local contrast measure (TLLCM)\cite{13}, improved local contrast measure (ILCM)\cite{14}, and relative local contrast measure (RLCM)\cite{6} assumed that the target's brightness was higher than its neighborhood, thereby failed to effectively detect dim targets. On the other hand, low-rank decomposition-based methods, including infrared patch-image (IPI)\cite{7}, non-convex rank approximation minimization joint $l_{2,1}$ norm (NRAM)\cite{8}, reweighted infrared patch-tensor (RIPT)\cite{15}, and partial sum of the tensor nuclear norm (PSTNN)\cite{16}, achieved target background separation based on the assumption of a low-rank background and sparse target, but they were susceptible to background clutter and lack strong adaptability. However, real-world scenes often exhibit a high level of background complexity, characterized by clutter and noise. Moreover, the target typically manifests as a faint feature due to the long imaging distance. Consequently, the performance of conventional methods is hindered by these limitations, leading to poor detection performance in real-world scenarios.
\subsection{Deep-learning based method}
Data-driven methods leveraging deep learning techniques demonstrated the ability to adaptively extract features from images and acquire high-level semantic information. Accordingly, the deep-learning based methods exhibited superior performance compared to traditional approaches when confronted with various complex environments. Moreover, with the opening of numerous infrared small target datasets attracted increasing interest among researchers on deep-learning based methods. Based on distinct processing paradigms, deep-learning based approaches can be categorized into two main groups: detection-based and segmentation-based methods.
\subsubsection{Segmentation-based methods}
Segmentation-based methods employed pixel-by-pixel threshold segmentation on the image, yielded a segmentation mask that provided object position and size information. Wang \emph{et al.}\cite{9} introduced a generative adversarial network (GAN) framework for adversarial learning, enabling the natural attainment of Nash equilibrium between miss detection (MD) and false alarm (FA) during training. Dai \emph{et al.}\cite{10} proposed an asymmetric contextual modulation (ACM) module that combined top-down and bottom-up point-wise attention mechanisms to enhance the encoding of semantic information and spatial details. Additionally, Dai \emph{et al.}\cite{17} presented a model-driven deep network attentional local contrast networks (ALCNet) that effectively utilized labeled data and domain knowledge, addressed issues such as inaccurate modeling, hyper-parameter sensitivity, and insufficient intrinsic features. Zhang \emph{et al.}\cite{18} introduced the infrared shape network (ISNet) for detecting shape information in infrared small target. To mitigate deep information loss caused by pooling layers in infrared small target, Li \emph{et al.}\cite{19} proposed the dense nested attention network (DNA-Net). Hou \emph{et al.}\cite{32} devised a robust infrared small target detection network (RISTDNet) that combined handcrafted feature methods with CNN. Chen \emph{et al.}\cite{20} developed a hierarchical overlapped small patch transformer (HOSPT) as a replacement for convolution kernels in convolutional neural network (CNN), enabled the encoding of multi-scale features and addressed the challenge of modeling long-range dependencies in images.
\subsubsection{Detection-based methods}
Detection-based methods were the same as the ordinary object detection algorithms, they directly outputed the target's position and scale information. To enhance the detection performance of infrared small target, Li \emph{et al.}\cite{21} proposed a method that incorporated super-resolution enhancement of the input image and improved the structure of YOLOv5. In a similar vein, Zhou \emph{et al.}\cite{22} tackled the challenge of detecting infrared small target by employing a YOLO-based framework. Dai \emph{et al.}\cite{23} introduced a one-stage cascade refinement network (OSCAR) to address the issues of inherent characteristics deficiency and inaccurate bounding box regression in infrared small target detection. Meanwhile, Yao \emph{et al.}\cite{24} developed a lightweight network that combined traditional filtering methods with the standard convolutional one stage object detection (FCOS) to improve responsiveness to infrared small target. Du \emph{et al.}\cite{25} adopted an interframe energy accumulation (IFEA) enhancement mechanism to amplify the energy of moving time series targets. Furthermore, the issue of sample misidentification was resolved by employing an small intersection over union (IoU) strategy. Similarly, Ju \emph{et al.}\cite{26} achieved the same objective through the utilization of an image filtering module.
\section{METHODOLOGY}\label{section3}
\subsection{Overall Architecture}
\begin{figure*}[htbp]
\centering
\includegraphics[trim= 0 0 0 0,clip,scale=0.4]{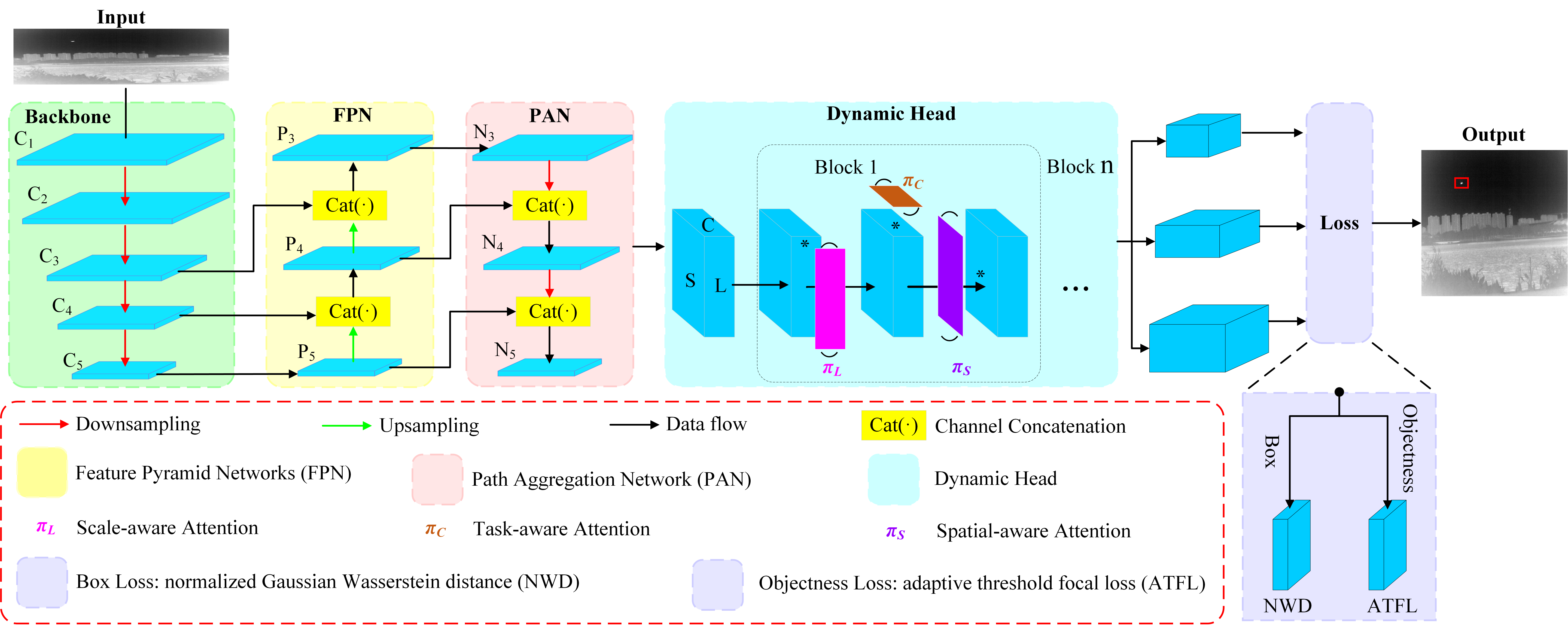}
\caption{ Overview of the proposed EFLNet, which has the structure of backbone, FPN, PAN, and dynamic head, as well as the loss functions of NWD and ATFL.}
\label{fig_1}
\end{figure*}

Fig. \ref{fig_1} shows the workflow of the proposed method. First, the infrared image serves as the input to the backbone network, enabling the extraction of essential features. These features undergo fusion via FPN and PAN \cite{33}, integrating multi-scale information. The resulting fused features are then fed into the dynamic detection head, facilitating the learning of the relative significance of diverse semantic layers. Ultimately, the detection results are assessed by the NWD and ATFL, which compute the loss and guide the model optimization process.
\subsection{Adaptive threshold focal loss}\label{section3-2}
\begin{figure}[H]
\centering
\includegraphics[width=3.5in]{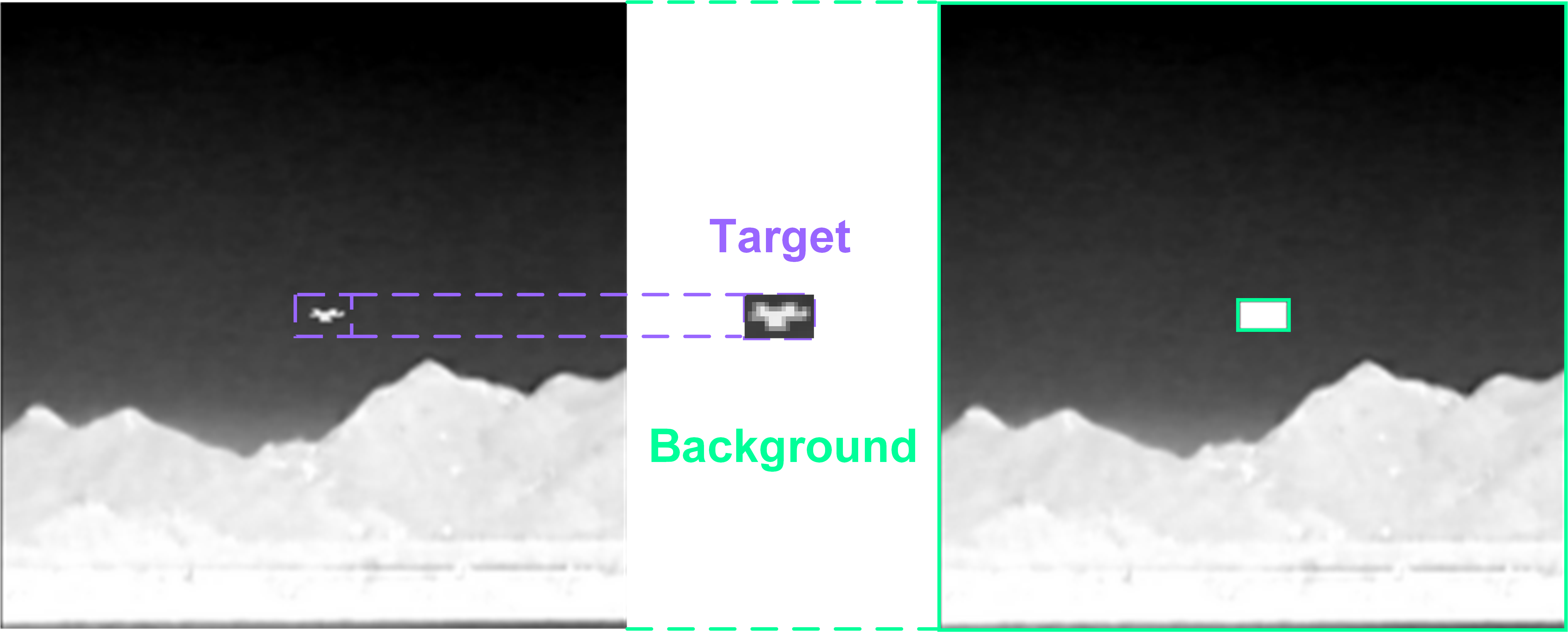}
\caption{The imbalance phenomenon between the target and the background.}
\label{fig_2}
\end{figure}

The infrared image predominantly consists of background, with only a small portion occupied by the target, as illustrated in Fig. \ref{fig_2}. Thus, learning the characteristics of the background during the training process is easier than learning those of the target.  The background can be considered as easy samples, while the targets can be regarded as hard samples. However, even the well-learned background still produces losses during training. In fact, the background samples that occupy the main part of the infrared image dominate the gradient update direction, overwhelmed the target information. To address this issue,we propose a new ATFL function. First, the threshold setting is used to decouple the easy-to-identify background from the difficult-to-identify target. Second, by intensifying the loss associated with the target and mitigating the loss linked to the background, we force the model to allocate greater attention to target features, thereby alleviating the imbalance between the target and the background. Finally, adaptive design has been applied to hyperparameters to reduce time consumption caused by adjusting hyperparameters.

We propose an ATFL that decouples the target and background based on the set threshold. The loss value is adaptively adjusted according to the predicted probability value, aiming to enhance the detection performance of infrared small target.

The classical cross-entropy loss function can be expressed as:
\begin{equation}
{{\cal L}_{{\rm{BCE}}}} =  - (y\log (p) + (1 - y)\log (1 - p))
\end{equation}
where $p$ represents the predicted probability and $y$ represents the true label. Its succinct representation is:
\begin{equation}
{{\cal L}_{{\rm{BCE}}}} =  - \log ({p_t})
\end{equation}
where 
\begin{equation}
p_{t}=\left\{\begin{array}{ll}
p, & \text { if } y=1 \\
1-p, & \text { others }
\end{array}\right.
\end{equation}
The cross-entropy function cannot address the imbalance problem between samples, so the focal loss\cite{27} function introduces a modulation factor $(1-p_t)^\gamma$ to reduce the loss contribution of easily classifiable samples by adjusting the focusing parameter $\gamma$. The focus loss function can be expressed as:
\begin{equation}
FL\left( {{p_{\rm{t}}}} \right) =  - {\left( {1 - {p_{\rm{t}}}} \right)^\gamma }\log \left( {{p_{\rm{t}}}} \right)
\end{equation}

\begin{figure}[htbp]
\centering
\includegraphics[width=2.5in]{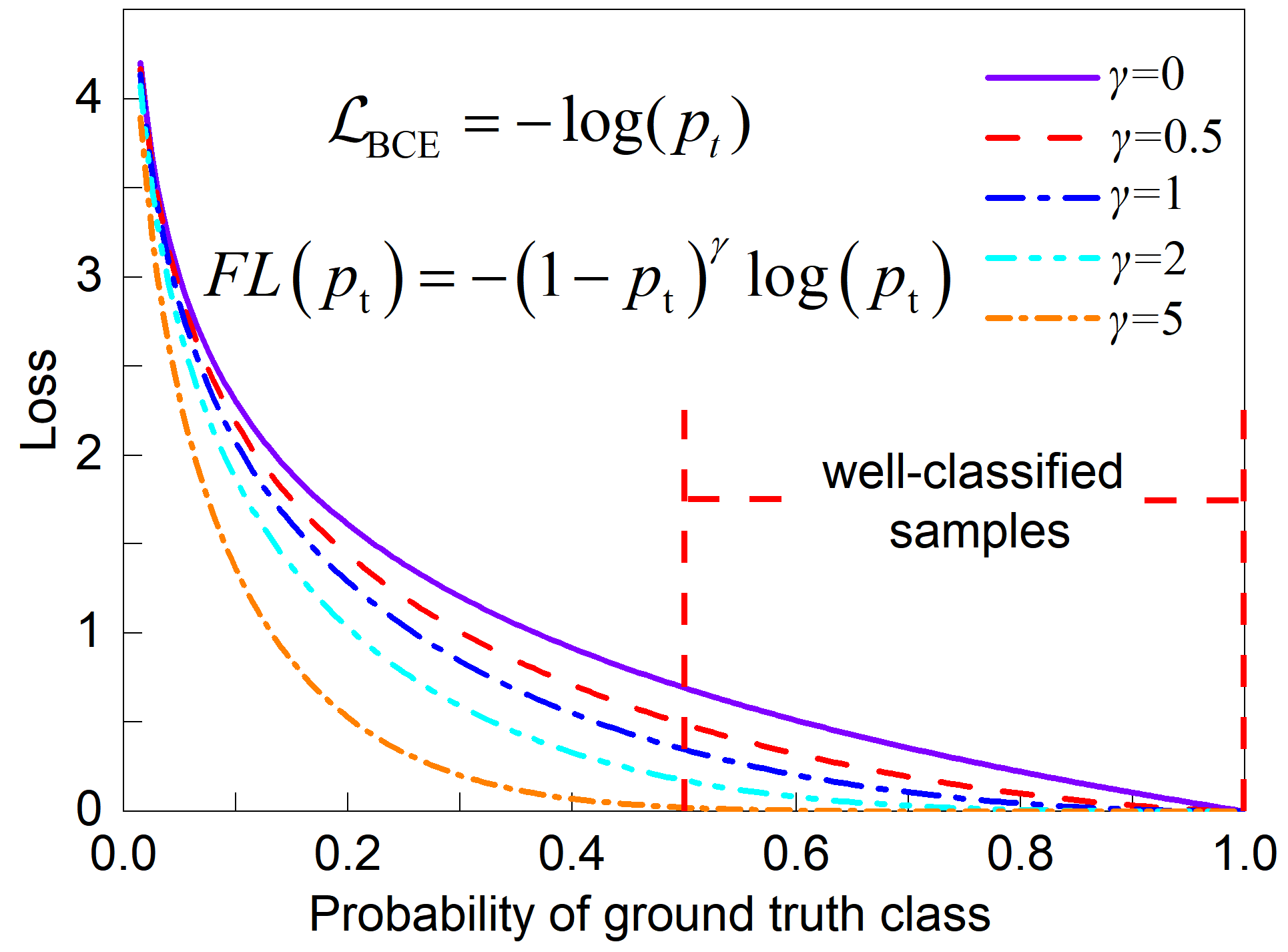}
\caption{Changes in losses in terms of different $\gamma$. The $p_t>0.5$ is regarded as well-classified samples.}
\label{fig_3}
\end{figure}

The focal loss function can adjust the value of the $\gamma$ to reduce the loss weight of easy samples, as can be seen in the Fig. \ref{fig_3}. However, while reducing the loss of easy samples, the modulation factor also reduces the value of difficult sample losses, which is not conducive to the learning of difficult samples.

To address above problem, we propose a threshold focal loss ($TFL$) function, which effectively mitigates the impact of easy samples by reducing their loss weight, while simultaneously increasing the loss weight assigned to difficult samples. Specifically, we designate prediction probability value above 0.5 as  easy samples, while conversely considering values below this threshold as hard samples. The expression is as follows:
\begin{equation}
\label{eq5}
T F L=\left\{\begin{array}{ll}
-\left(\lambda-p_{t}\right)^{\eta} \log \left(p_{\mathrm{t}}\right) & p_{\mathrm{t}}<=0.5 \\
-\left(1-p_{t}\right)^{\gamma} \log \left(p_{\mathrm{t}}\right) & p_{\mathrm{t}}>0.5
\end{array}\right.
\end{equation}
where $\eta$, $\gamma$, $\lambda(>1)$ are the hyperparameters. For different datasets and models, the hyperparameters need to be adjusted multiple times to achieve optimal performance. In the field of artificial intelligence, each training takes a lot of time, resulting in expensive time costs. Therefore, we have made adaptive improvements to $\eta$ and $\gamma$.

For easy samples, we expect the loss value to decrease as $p_t$ increases, further reducing the loss generated by easy samples. At the beginning of training, even easy samples will have a relatively low prediction probability and gradually rise as the training process progresses, and $\gamma$ should gradually approach 0. The predicted probability value $\hat p_c$ of the real target can be used to mathematically model the progress of model training, and it can be predicted by exponential smoothing. It is stated as follows:
\begin{equation}
{\hat p_c} = 0.05 \times \frac{1}{{t - 1}}\sum\limits_{i = 0}^{t - 1} {\overline {{p_i}} }  + 0.95 \times {p_t}
\end{equation}
where $\hat p_c$  represents the predicted value for the next epoch, $p_t$ represents the current average predicted probability value, and $\overline p_i$ represents the average predicted probability value for each training epoch. According to Shannon's information theory, the greater the probability value of an event, the smaller the amount of information it brings; Conversely, the greater the amount of information. Thus, the adaptive modulation factor $\gamma$ can be expressed as:
\begin{equation}
\label{eq7}
\gamma  =  - \ln \left( {{{\hat p}_{\rm{c}}}} \right)
\end{equation}
However, in the later stage of network training, the expected probability value is too large, which will reduce the proportion of difficult samples. We express the $\eta$ as:
\begin{equation}
\label{eq8}
\eta  =  - \ln ({p_t})
\end{equation}
By incorporating Eq.(\ref{eq7}), (\ref{eq8}) into Eq.(\ref{eq5}), the expression of the adaptive threshold focal loss can be obtained as:
\begin{equation}
A T F L= \begin{cases}-\left(\lambda-p_t\right)^{-\ln \left(p_t\right)} \log \left(p_{\mathrm{t}}\right) & p_{\mathrm{t}}<=0.5 \\ -\left(1-p_t\right)^{-\ln \left(\hat{p}_c\right)} \log \left(p_{\mathrm{t}}\right) & p_{\mathrm{t}}>0.5\end{cases}
\end{equation}
\subsection{Normalized Gaussian Wasserstein distance}\label{section3-3}
\begin{figure*}[htbp]
\centering
\subfloat[]{\includegraphics[width=3in]{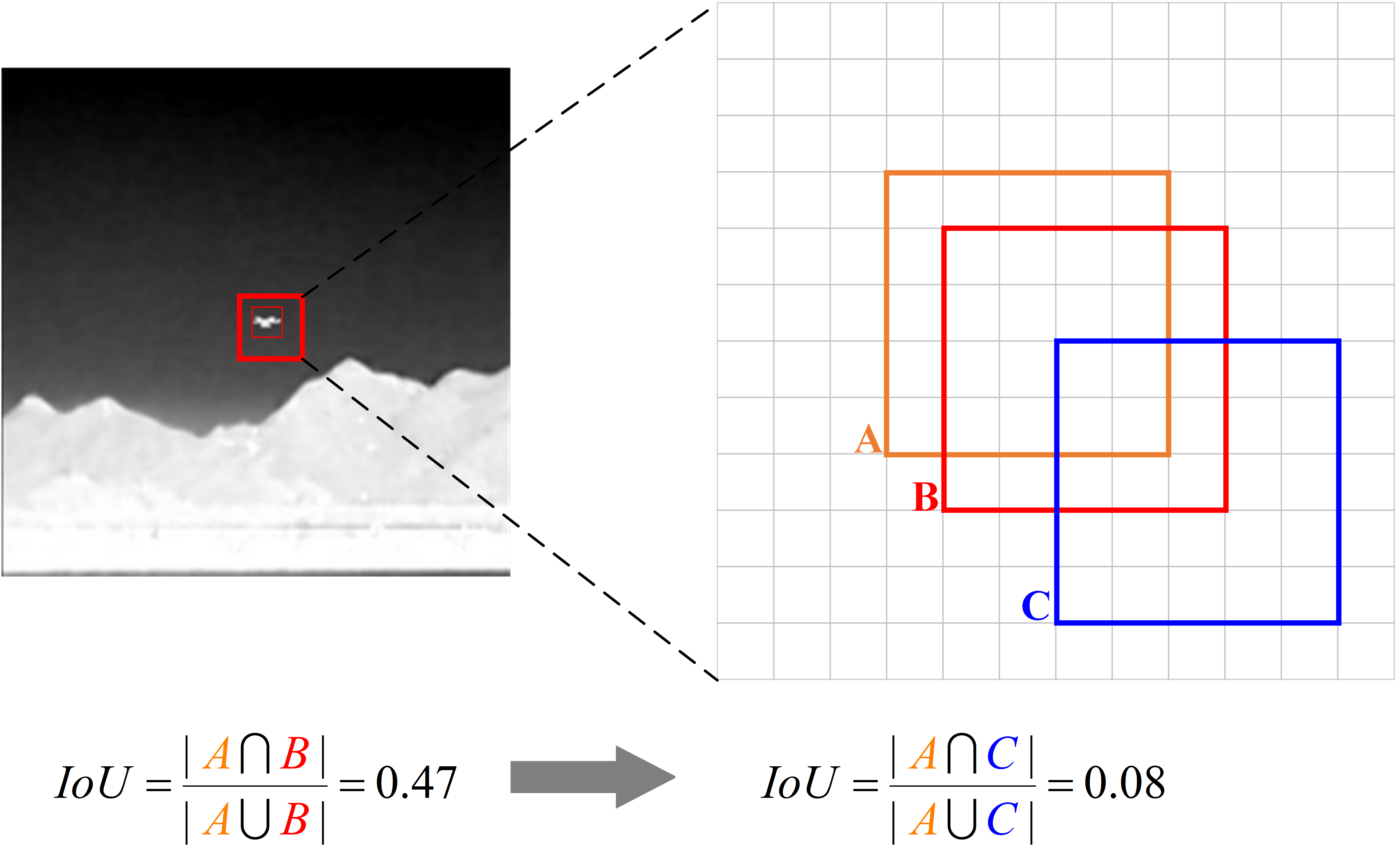}%
\label{tiny scale object}}
\hfil
\subfloat[]{\includegraphics[width=3in]{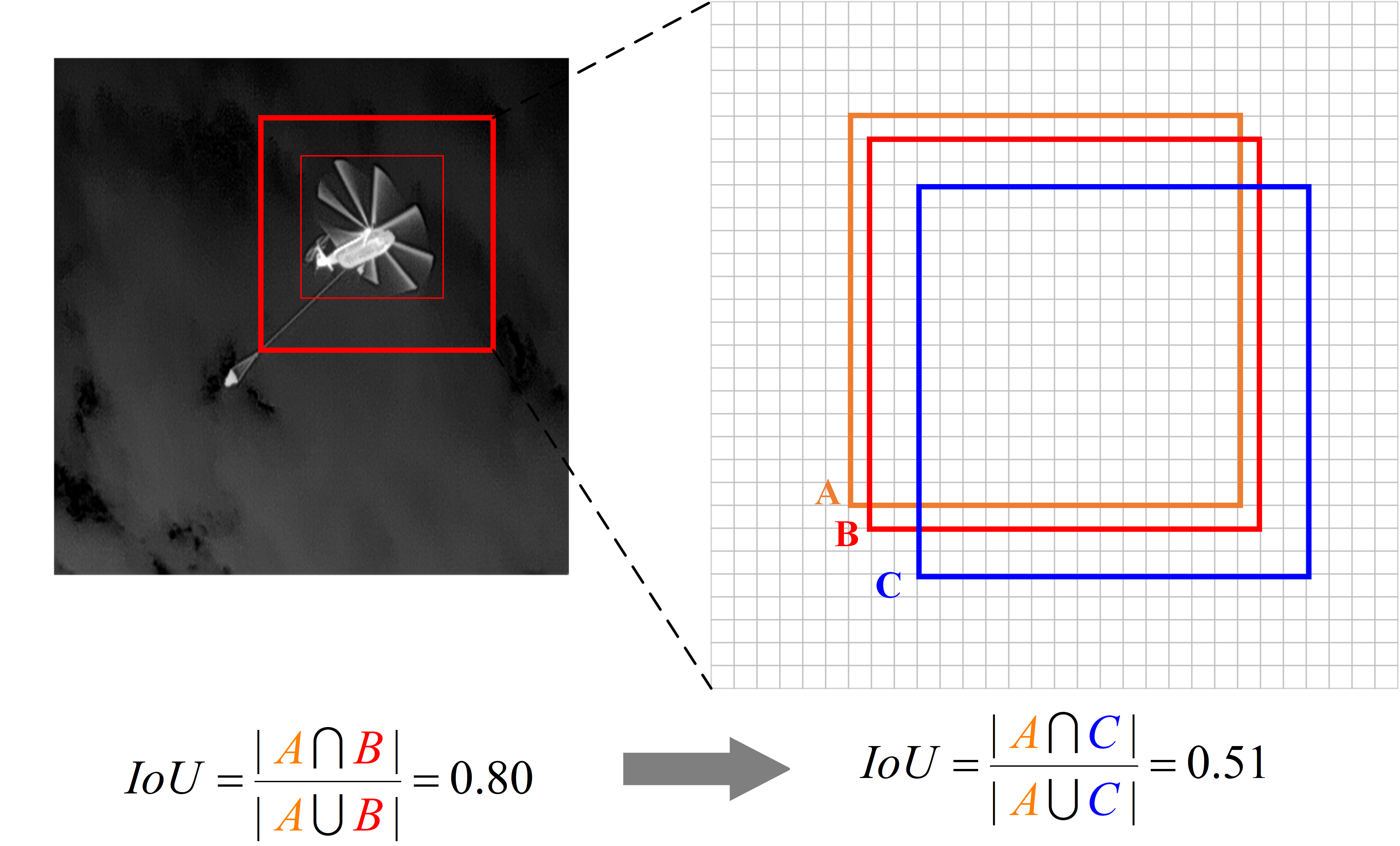}%
\label{ normal scale object}}
\caption{Sensitivity analysis of IoU on tiny and normal scale object. (a) Tiny scale object. (b) Normal scale object.}
\label{fig_4}
\end{figure*}

The IoU metric used for ordinary object detection exhibits extreme sensitivity when applied to infrared small target. Even a slight deviation in position between the predicted boxes and ground-truth boxes can result in a significant change in IoU. This sensitivity is illustrated in the Fig. \ref{fig_4}, where a small position deviation leads to a decrease in IoU for infrared small target from 0.47 to 0.08. Conversely, for normal-sized objects, the IoU only decreases from 0.80 to 0.51 under the same position deviation. Such sensitivity of the IoU metric towards infrared small target leads to a high degree of similarity between positive and negative samples during training, making it challenging for the network to converge effectively. Furthermore, in extreme cases, the infrared small target may occupy only one or a few pixels within the image. Consequently, the IoU between the ground truth and any predicted bounding box falls below the minimum threshold, resulting in zero positive samples within the image. Therefore, alternative evaluation indicators are required for assessing infrared small target more accurately.

The IoU metric is actually a similarity calculation between samples, which is sensitive to the size change of the target and is not suitable for infrared small target, so we introduce NWD as a new measure. The Wasserstein distance can measure the similarity between distributions with minimal or no overlap, and it is also insensitive to objects of different scales. Therefore, it can address issues related to the similarity of positive and negative samples, as well as sparse positive samples during the training process of infrared small target. Specifically, the bounding box is modeled as a 2D Gaussian distribution:
\begin{equation}
f(\mathbf{x} \mid \boldsymbol{\mu}, \boldsymbol{\Sigma})=\frac{\exp \left(-\frac{1}{2}(\mathbf{x}-\boldsymbol{\mu})^{\top} \boldsymbol{\Sigma}^{-1}(\mathbf{x}-\boldsymbol{\mu})\right)}{2 \pi|\mathbf{\Sigma}|^{\frac{1}{2}}}
\end{equation}
where $\mathbf{x}$, $\boldsymbol{\mu}$ and $\boldsymbol{\Sigma}$ represent the coordinates($x$, $y$), the mean vector, and co-variance matrix of the Gaussian distribution. When
\begin{equation}
(\mathbf{x}-\boldsymbol{\mu})^{\top} \boldsymbol{\Sigma}^{-1}(\mathbf{x}-\boldsymbol{\mu})=1
\end{equation}
The horizontal bounding box $R=(c_x, c_y, w, h)$ can be modeled as a 2D Gaussian distribution using $N(\boldsymbol{\mu}, \boldsymbol{\Sigma})$:
\begin{equation}
{\bf{\mu }} = \left[ \begin{array}{l}
{c_x}\\
{c_y}
\end{array} \right],{\rm{\quad}}\boldsymbol{\Sigma} {{\bf{ = }}\left[ \begin{array}{l}
\frac{{{w^2}}}{4}{\bf{\quad }}0\\
{\bf{ }}0{\bf{\quad }}\frac{{{h^2}}}{4}
\end{array} \right]} 
\end{equation} 
where $(c_x, c_y)$, $w$ and $h$ represent the center coordinates, width and height, respectively. The 2D Wasserstein distance between two 2D Gaussian distributions $\mu_1=N(\boldsymbol{m_1}, \boldsymbol{\Sigma}_1)$ amd $\mu_2=N(\boldsymbol{m_2}, \boldsymbol{\Sigma}_2)$ is defined as:
\begin{equation}
\begin{array}{l}
W_2^2\left( {{\mu _1},{\mu _2}} \right) = \parallel {{\bf{m}}_1} - {{\bf{m}}_2}\parallel _2^2 +\\
{\rm{\quad\quad\quad\quad\quad\quad}}{\bf{Tr}}\left( {{{\bf{\Sigma }}_1} + {{ \boldsymbol{\Sigma}}_2} - 2{{\left( {{ \boldsymbol{\Sigma}}_2^{1/2}{{ \boldsymbol{\Sigma}}_1}{\boldsymbol{\Sigma}}_2^{1/2}} \right)}^{1/2}}} \right)
\end{array}
\end{equation}
It can be simplified as:
\begin{equation}
W_2^2\left( {{\mu _1},{\mu _2}} \right) = \parallel {{\bf{m}}_1} - {{\bf{m}}_2}\parallel _2^2 + \parallel {\boldsymbol{\Sigma}}_1^{1/2} - {\boldsymbol{\Sigma}}_2^{1/2}\parallel _F^2
\end{equation}
where $\parallel\cdot\parallel_F $ is the Frobenius norm. The distance between the Gaussian distributions $N_a$, $N_b$ modeled by bounding boxes $A=(cx_a, cy_a, w_a, h_a)$ and $B=(cx_b, cy_b, w_b, h_b)$ can be simplified as:
\begin{equation}
W_2^2\left( {{{\cal N}_a},{{\cal N}_b}} \right) = \left( {{{\left[ {c{x_a},c{y_a},\frac{{{w_a}}}{2},\frac{{{h_a}}}{2}} \right]}^{\rm{T}}},{{\left[ {c{x_b},c{y_b},\frac{{{w_b}}}{2},\frac{{{h_b}}}{2}} \right]}^{\rm{T}}}} \right)_2^2
\end{equation}
Normalizing it exponentially to a range of 0-1 gives the normalized Watherstein distance\cite{28}:
\begin{equation}
NWD\left( {{{\cal N}_a},{{\cal N}_b}} \right) = \exp \left( { - \frac{{\sqrt {W_2^2\left( {{{\cal N}_a},{{\cal N}_b}} \right)} }}{C}} \right)
\end{equation}
where $C$ is a constant related to the dataset.
\subsection{Dynamic head}\label{section3-4}
\begin{figure}[htbp]
\centering
\includegraphics[width=3.6in]{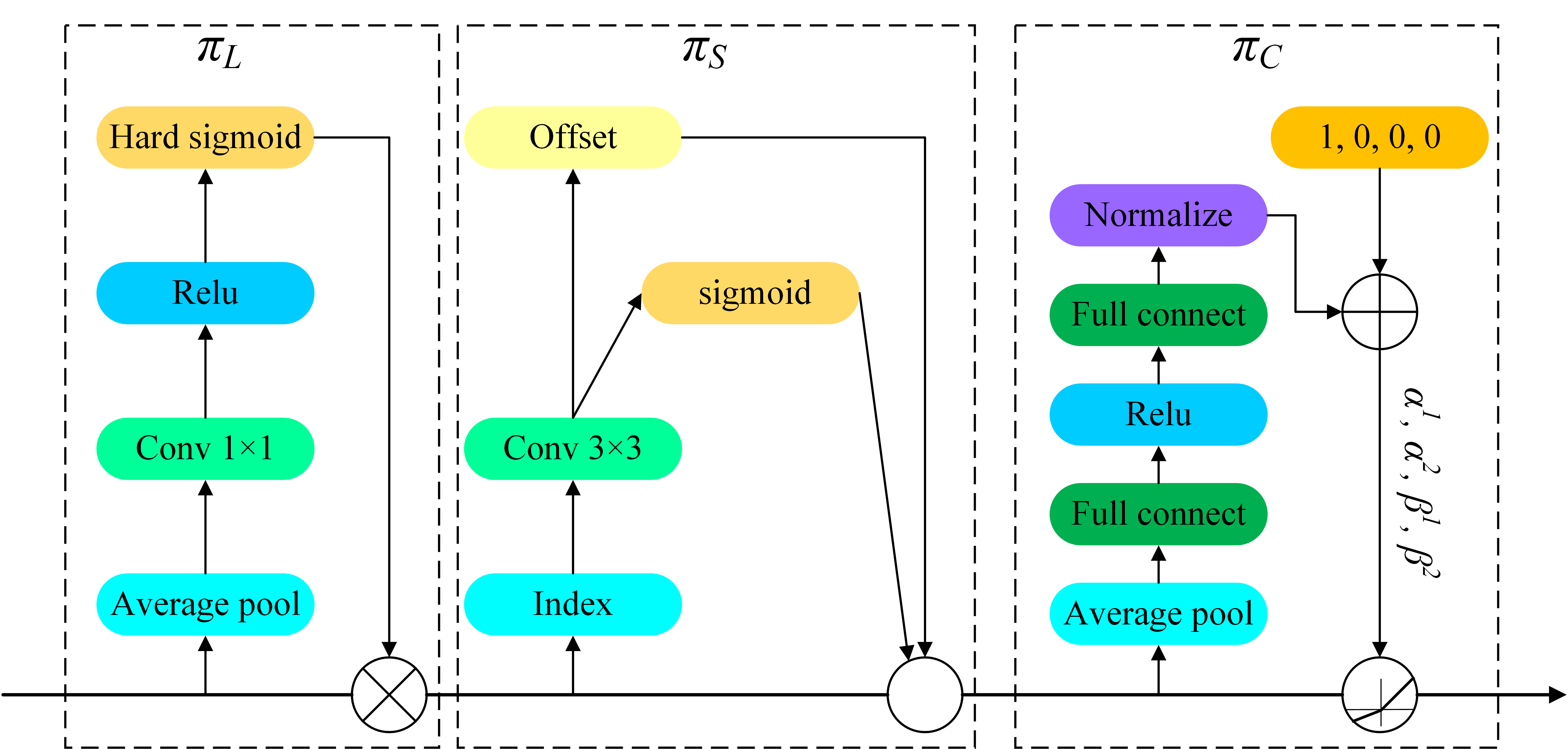}
\caption{Structure of dynamic head block. The $\pi_L$ denotes scale-aware attention, $\pi _S$ is spatial-aware attention, and $\pi _C$ represents task-aware attention.}
\label{fig_5}
\end{figure}

Feature pyramid networks, which involve combining multi-scale convolution features, have become a prevalent technique in detection networks. Nevertheless, it is important to note that features at varying depths convey distinct semantic information. Specifically, during the down-sampling process, infrared small target may experience information loss. Shallow features contain valuable infrared small target information that merits greater attention from the network. On the other hand, different viewpoints and task forms will produce different features and target constraints, which bring difficulties to infrared small target detection. The dynamic head as shown in Fig. \ref{fig_5} can adaptively focus on the scale-space-task information of objects, which can better learn the relative importance of each semantic level and spatial information of the target, as well as adaptively match different task forms. 

Given the feature tensor $F\in R^{L\times S \times C}$, $L$ represents the number of pyramid layers, $S$ represents the size of the feature,$S=H\times W$ , $H$, $W$ represents the height and width of the feature, and $C$ represents the number of channels. Dynamic head \cite{29} can be expressed as:
\begin{equation}
W({\cal F}) = {\pi _C}\left( {{\pi _S}\left( {{\pi _L}({\cal F}) \cdot {\cal F}} \right) \cdot {\cal F}} \right) \cdot {\cal F}
\end{equation}
where $\pi_L(\cdot)$, $\pi_S(\cdot)$, $\pi_C(\cdot)$ represents the attention function on $L$, $S$, and $C$, respectively. Scale-aware attention $\pi_L$ enables dynamic feature fusion based on the importance of features in each layer:
\begin{equation}
{\pi _L}({\cal F}) \cdot {\cal F} = \sigma \left( {f\left( {\frac{1}{{SC}}\mathop \sum \limits_{S,C} {\cal F}} \right)} \right) \cdot {\cal F}
\end{equation}
where $f(\cdot)$ is a $1\times1$ convolutional layer $\sigma(x)=max(0,min(1,\frac{x+1}{2}))$ is a hard-sigmoid function. 

Spatial-aware attention $\pi_S(\cdot)$ uses deformable convolution\cite{30} to fuse features of different levels in the same spatial position.
\begin{equation}
{\pi _S}({\cal F}) \cdot {\cal F} = \frac{1}{L}\mathop \sum \limits_{l = 1}^L \mathop \sum \limits_{k = 1}^K {w_{l,k}} \cdot {\cal F}\left( {l;{p_k} + {\rm{\Delta }}{p_k};c} \right) \cdot {\rm{\Delta }}{m_k}
\end{equation}
where $K$ is the number of sparse sampling locations, $p_k+\Delta p_k$ and $\Delta m_k$ are learned from the input features, $p_k+\Delta p_k$ is a shifted location by the self-learned spatial offset $\Delta p_k$, $\Delta m_k$ is an important scalar for self-learning at position $p_k$. 
Task-aware attention $\pi_C(\cdot)$ dynamically switches $ON$ and $OFF$ channels to support different tasks: 
\begin{equation}
{\pi _C}({\cal F}) \cdot {\cal F} = \max \left( {{\alpha ^1}({\cal F}) \cdot {{\cal F}_c} + {\beta ^1}({\cal F}),{\alpha ^2}({\cal F}) \cdot {{\cal F}_c} + {\beta ^2}({\cal F})} \right)
\end{equation}
where $[\alpha^1,\alpha^2,\beta^1,\beta^2]^T$ is a superfunction that controls the threshold, which reduces the dimension in the $L\times S$ dimension through average pooling, then uses two fully connected layers and a normalization layer, and finally normalizes by the sigmoid activation function.
\section{Experiment}\label{section4}
\subsection{Dataset}
\begin{figure}[htbp]
\centering
\subfloat[]{
\begin{minipage}[t]{0.31\linewidth}
\centering
\includegraphics[width=2.8cm,height=2.8cm]{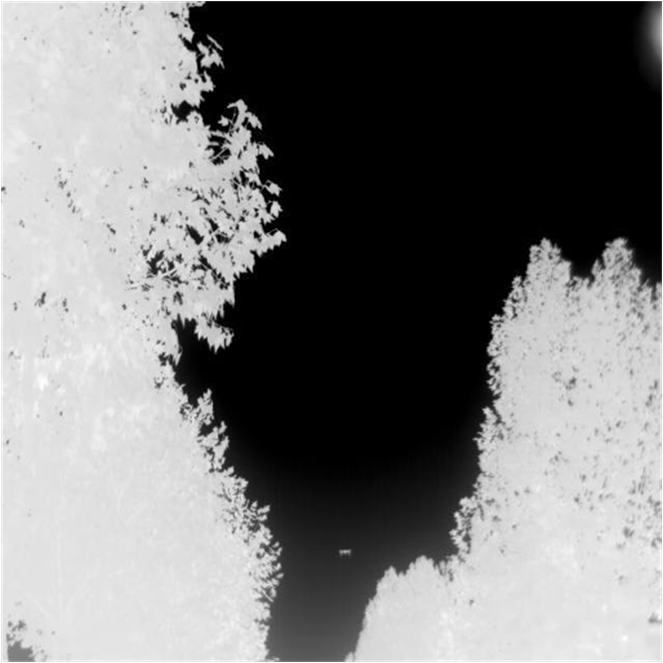}
\end{minipage}%
}%
\subfloat[]{
\begin{minipage}[t]{0.31\linewidth}
\centering
\includegraphics[width=2.8cm,height=2.8cm]{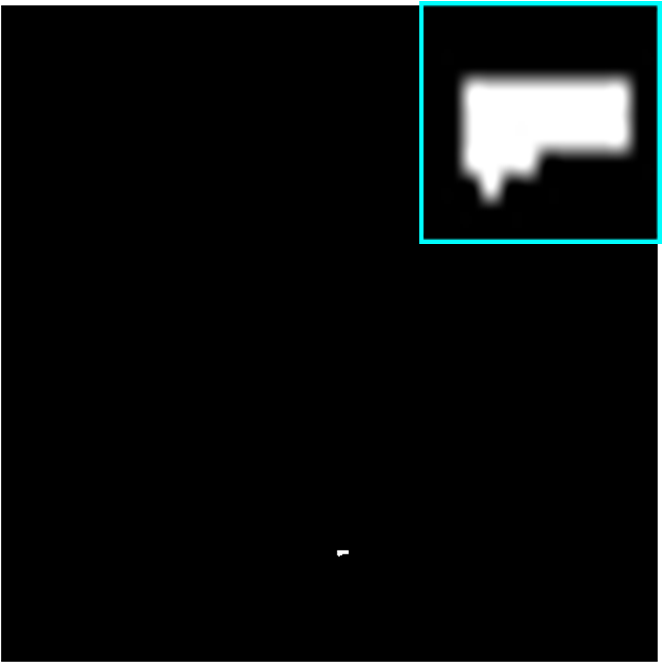}
\end{minipage}%
}%
\subfloat[]{
\begin{minipage}[t]{0.31\linewidth}
\centering
\includegraphics[width=2.8cm,height=2.8cm]{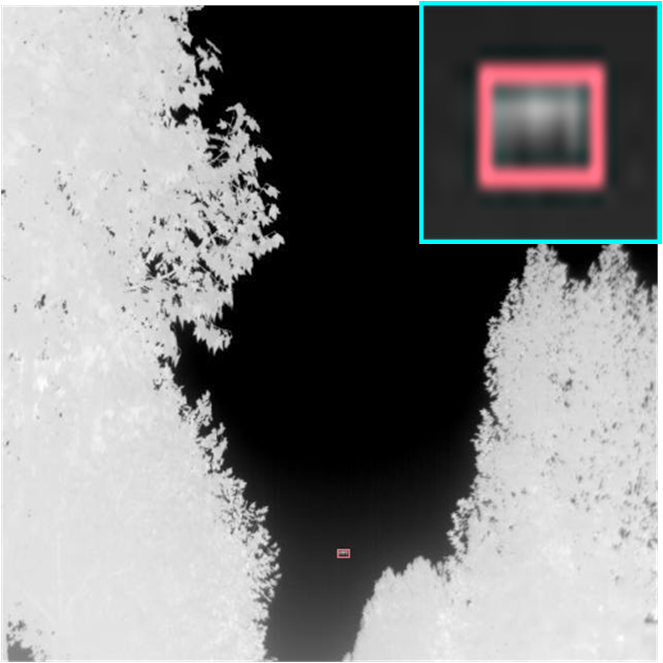}
\end{minipage}%
}%
\centering
\caption{The different annotation forms for the current infrared small target public dataset. (a) Image. (b) Semantic segmentation. (c) Bounding box.}
\label{fig_7_1}
\end{figure}

\textbf{Datasets:} We conducted experiments using bounding box annotation and semantic segmentation mask annotation from three publicly available infrared small target datasets (as depicted in Fig. \ref{fig_7_1}): NUAA-SIRST\cite{10}, NUDT-SIRST\cite{19}, and IRSTD-1k\cite{18}. To ensure proper evaluation, we divided each dataset into training set, validation set and test set, following a ratio of 6:2:2.

\textbf{Evaluation metrics:} To compare the proposed method with the state-of-the-art (SOTA) methods, we employ commonly used evaluation metrics including precision, recall and F1. Each metric is defined as follows:

$Precision$: Precision is calculated as the ratio of true positives (TP) to the sum of true positives and false positives (FP):
\begin{equation}
Precision = TP/(TP + FP)
\end{equation}

$Recall$: Recall is calculated as the ratio of true positives to the sum of true positives and false negatives (FN):
\begin{equation}
Recall = TP/(TP + FN)
\end{equation}

$F1$: F1 is a harmonic mean of precision and recall, providing a balanced measure of the model's performance, which is computed as:
\begin{equation}
F1 = 2*(Precision*Racll)/(Precision + Racll)
\end{equation}

\textbf{Comparison to the SOTA Methods:} Deep-learning methods have exhibited significantly superior performance compared to model-based methods, such as Top-Hat, Max-Median, WSLCM, TLLCM, NRAM, IPI, RIPT. These results have been widely demonstrated\cite{9,17,18,19}. Therefore, this paper does not compare with model-based methods anymore, but with several deep-learning based most advanced methods, including MDvsFA\cite{9}, AGPCNet\cite{31}, ACM\cite{10}, ISNet\cite{18}, ALCNet\cite{17}, DNANet\cite{19}. To ensure fair comparisons, each model has been retrained using a repartitioned dataset, employing a training epoch of 400 and keeping the remaining parameters at their default values.
\subsection{Quantitative Results}
Table \ref{table1} presents a quantitative comparison of the results obtained from different methods. The proposed method demonstrates the highest performance across all evaluation metrics on the NUAA-SIRST, NUDT-SIRST and IRSTD-1k datasets when compared to the SOTA method which proves the effectiveness of the proposed method. Because most of the current deep-learning based methods regard infrared small target detection as a pixel-level segmentation task, pixel-level segmentation results need to be obtained. The slightest inadequacy will produce false alarms or missed detections, resulting in poor detection performance at the target level. In addition, there is less attention paid to the imbalance phenomenon and boundary box sensitivity issues in infrared small target. Therefore, these methods often yield poor performance in target-level detection tasks, resulting in relatively low precision, recall, and F1 scores. We designed ATFL to solve the imbalance between target and background by adaptive adjustment of loss weight, and make the model better learn the features of infrared small target through NWD metric and dynamic head. Therefore, it shows better detection performance for infrared small target.
\begin{table*}[htbp]
\caption{Comparisons with SOTA methods on NUAA-SIRST, NUDT-SIRST and IRSTD-1k in precision, recall and F1.\label{tab:table1}}
\centering
\begin{tabular}{|c|ccc|ccc|ccc|}
\hline
\multirow{2}{*}{Method} & \multicolumn{3}{c|}{NUAA-SIRST}                                 & \multicolumn{3}{c|}{NUDT-SIRST}                                 & \multicolumn{3}{c|}{IRSTD-1k}                                   \\ \cline{2-10} 
                        & \multicolumn{1}{c|}{$Precision$}   & \multicolumn{1}{c|}{$Recall$}   & $F1$    & \multicolumn{1}{c|}{$Precision$}   & \multicolumn{1}{c|}{$Recall$}   & $F1$    & \multicolumn{1}{c|}{$Precision$}   & \multicolumn{1}{c|}{$Recall$}   & $F1$    \\ \hline
MDvsFA[\textcolor{green}{9}]                  & \multicolumn{1}{c|}{0.845} & \multicolumn{1}{c|}{0.507} & 0.597 & \multicolumn{1}{c|}{0.608} & \multicolumn{1}{c|}{0.192} & 0.262 & \multicolumn{1}{c|}{0.550}  & \multicolumn{1}{c|}{0.483} & 0.475 \\ \hline
AGPCNet[\textcolor{green}{31}]                & \multicolumn{1}{c|}{0.390}  & \multicolumn{1}{c|}{0.810}  & 0.527 & \multicolumn{1}{c|}{0.368} & \multicolumn{1}{c|}{0.684} & 0.479 & \multicolumn{1}{c|}{0.415} & \multicolumn{1}{c|}{0.470}  & 0.441 \\ \hline
ACM[\textcolor{green}{10}]                     & \multicolumn{1}{c|}{0.765} & \multicolumn{1}{c|}{0.762} & 0.763 & \multicolumn{1}{c|}{0.732} & \multicolumn{1}{c|}{0.745} & 0.738 & \multicolumn{1}{c|}{0.679} & \multicolumn{1}{c|}{0.605} & 0.640  \\ \hline
ISNet[\textcolor{green}{18}]                  & \multicolumn{1}{c|}{0.820}  & \multicolumn{1}{c|}{0.847} & 0.834 & \multicolumn{1}{c|}{0.742} & \multicolumn{1}{c|}{0.834} & 0.785 & \multicolumn{1}{c|}{0.718} & \multicolumn{1}{c|}{0.741} & 0.729 \\ \hline
ACLNet[\textcolor{green}{17}]                  & \multicolumn{1}{c|}{0.848} & \multicolumn{1}{c|}{0.78}  & 0.813 & \multicolumn{1}{c|}{0.868} & \multicolumn{1}{c|}{0.772} & 0.817 & \multicolumn{1}{c|}{0.843} & \multicolumn{1}{c|}{0.656} & 0.738 \\ \hline
DNANet[\textcolor{green}{19}]               & \multicolumn{1}{c|}{0.847} & \multicolumn{1}{c|}{0.836} & 0.841 & \multicolumn{1}{c|}{0.914} & \multicolumn{1}{c|}{0.889} & 0.901 & \multicolumn{1}{c|}{0.768} & \multicolumn{1}{c|}{0.721} & 0.744 \\ \hline
Ours                    & \multicolumn{1}{c|}{\textbf{0.882}} & \multicolumn{1}{c|}{\textbf{0.858}} & \textbf{0.870} & \multicolumn{1}{c|}{\textbf{0.963}} & \multicolumn{1}{c|}{\textbf{0.931}} & \textbf{0.947} & \multicolumn{1}{c|}{\textbf{0.870}} & \multicolumn{1}{c|}{\textbf{0.817}} & \textbf{0.843} \\ \hline
\end{tabular}
\label{table1}
\end{table*}
\subsection{Visual Results}
\begin{figure*}[hbp]
\centering
\includegraphics[trim= 0 0 0 0,clip,scale=0.35]{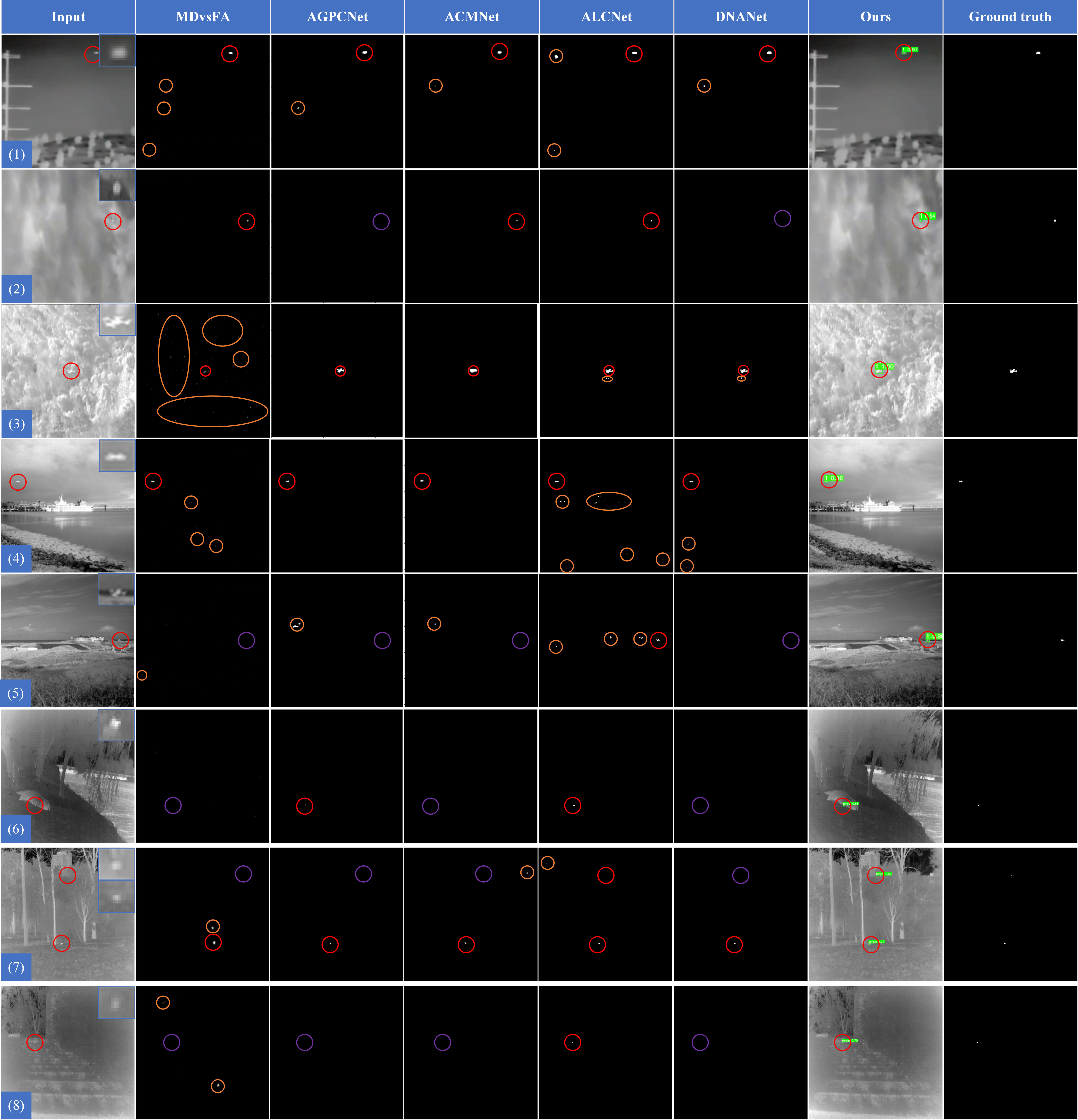}
\caption{Partial visual results gained by different methods on NUAA-SIRST, NUDT-SIRST and IRSTD-1k datasets. The targets are represented by circles colored in red, purple, and orange, indicating correctly detected targets, miss detected targets, and false detected targets, respectively.}
\label{fig_6}
\end{figure*}

The partial visualization results of different methods on the NUAA-SIRST, NUDT-SIRST and IRSTD-1k datasets are shown in Fig. \ref{fig_6}. The areas corresponding to correctly detected targets, false alarms, and missed detections are highlighted by circles in red, orange, and purple, respectively. The red circle indicates the correct target, the orange circle indicates the false alarm, and the purple circle indicates the missed detection. A model with superior detection performance exhibits a greater number of red circles and fewer orange and purple circles in the graph.  Generally, deep learning-based methods exhibit robust performance owing to their adaptive learning features, enabling effective detection of the majority of targets. However, most of these methods tend to produce false alarms when encountering locally highlighted interference (as shown in Fig. \ref{fig_6} (1), (4), (5)). Additionally, missed detection occurs when the target appears dim (as depicted in Fig. \ref{fig_6} (5), (7), (8)). Our proposed method effectively learns the characteristics of infrared small target, allowing for accurate detection and localization even in presence of local highlight interference and dim targets.
\subsection{Ablation Study}
We selected the IRSTD-1k dataset as our experimental dataset due to its composition of real images and an ample quantity of data. The NUAA dataset contains a limited number of images, while the NUDT dataset consists of images generated through simulation. To assess the effectiveness of each component within the EFLNet, we conducted multiple ablation experiments on the IRSTD-1k dataset.

\textbf{Impact of ATFL:} We investigated the effects of different hyperparameter forms and different values of $\lambda$  on ATFL. As shown in Table \ref{table2-0}, when using a fixed hyperparameter form, multiple adjustments to $\eta$  and $\gamma$ are required, which can be time-consuming. On the contrary, when employing adaptive hyperparameter, the optimized results can be obtained with a single tuning operation, eliminating the need for multiple parameter adjustments. 
\begin{table*}[htbp]
\centering
\caption{ Ablation study on the different hyperparameter form  of ATFL in precision, recall,  $AP_{0.5}$ and F1.\label{tab:table2}}
\begin{tabular}{|c|c|c|c|c|c|c|}
\hline
Hyperparameter form                 & $\eta$  & $\gamma$   & $Precision$   & $Recall$   & $AP_{0.5}$   & $F1$ \\ \hline
\multirow{10}{*}{Fixed hyperparameters} & 2  & 2  & 0.875                 & 0.723                 & 0.762                 & 0.792                 \\ \cline{2-7} 
                                        & 2  & 4  & 0.777                 & 0.760                 & 0.718                 & 0.768                 \\ \cline{2-7} 
                                        & 2  & 6  & 0.780                 & 0.762                 & 0.728                 & 0.771                 \\ \cline{2-7} 
                                        & 2  & 8  & 0.742                 & 0.804                 & 0.735                 & 0.772                 \\ \cline{2-7} 
                                        & 2  & 10 & 0.736                 & 0.727                 & 0.702                 & 0.731                 \\ \cline{2-7} 
                                        & 2  & 2  & 0.875                 & 0.723                 & 0.762                 & 0.792                 \\ \cline{2-7} 
                                        & 4  & 2  & 0.889                 & 0.698                 & 0.767                 & 0.782                 \\ \cline{2-7} 
                                        & 6  & 2  & 0.813                 & 0.712                 & 0.739                 & 0.759                 \\ \cline{2-7} 
                                        & 8  & 2  & 0.816                 & 0.669                 & 0.715                 & 0.735                 \\ \cline{2-7} 
                                        & 10 & 2  & 0.679                 & 0.756                 & 0.722                 & 0.715                 \\ \hline
Adaptive hyperparameters                & /  & /  & 0.876                 & 0.749                 & 0.780                 & 0.808                 \\ \hline
\end{tabular}
\label{table2-0}
\end{table*}
Furthermore, the adaptive mechanism yielded superior results compared to fixed hyperparameter. Thus, the effectiveness of our designed adaptive mechanism has been validated. As shown in Table \ref{table2}, we change the $\lambda$ values (e.g., 1.5, 2, 2.5, 3, 3.5, 4) and compared their impact on model performance against the baseline. The initial baseline model exhibits a relatively low detection rate for targets (recall=0.743). However, with the incorporation of the ATFL, the performance of model undergoes a significant enhancement. By assigning greater importance to hard-to-detect targets, the detection rate of infrared small target is improved, resulting in an enhanced recall rate of up to 0.790. Notably, when $\lambda=3.5$, the overall performance reaches its optimal level, validating the effectiveness of our method. As can be seen from the Table \ref{table2-1} that overly small or large thresholds result in decreased performance. This can be attributed to the imprecise classification of samples when the threshold is set unreasonable, potentially leading to negative effect. The optimal performance is achieved when the threshold is set at 0.5.
\begin{table}[H]
\centering
\caption{ Ablation study on the different parameter $\lambda$ of ATFL in precision, recall, $AP_{0.5}$ and F1.\label{tab:table2}}
\begin{tabular}{|c|c|c|c|c|}
\hline
\multicolumn{1}{|c|}{$\lambda$} & $Precision$   & $Recall$   & $AP_{0.5}$   & $F1$    \\ \hline
Baseline               & 0.845 & 0.756 & 0.770 & 0.798 \\ \hline
1.5                    & 0.851 & 0.768 & 0.773 & 0.807 \\ \hline
2                      & 0.879 & 0.726 & 0.763 & 0.795 \\ \hline
2.5                    & 0.876 & 0.749 & 0.780  & 0.808 \\ \hline
3                      & 0.889 & 0.745 & 0.763 & 0.810 \\ \hline
3.5                    & 0.851 & 0.790 & 0.790 & 0.819 \\ \hline
4                      & 0.870 & 0.736 & 0.748 & 0.797 \\ \hline
\end{tabular}
\label{table2}
\end{table}

\begin{table}[H]
\centering
\caption{ Ablation study on the different threshold setting of ATFL in precision, recall, $AP_{0.5}$ and F1.\label{tab:table2}}
\begin{tabular}{|c|c|c|c|c|}
\hline
Threshold   setting & $Precision$   & $Recall$   & $AP_{0.5}$   & $F1$     \\ \hline
Baseline            & 0.845 & 0.756 & 0.770 & 0.798 \\ \hline
0.1                 & 0.869 & 0.723 & 0.766 & 0.789 \\ \hline
0.3                 & 0.816 & 0.772 & 0.782 & 0.793 \\ \hline
0.5                 & 0.885 & 0.746 & 0.784 & 0.810 \\ \hline
0.7                 & 0.895 & 0.736 & 0.778 & 0.808 \\ \hline
0.9                 & 0.855 & 0.720 & 0.757 & 0.782 \\ \hline
\end{tabular}
\label{table2-1}
\end{table}

\begin{figure*}[]
\centering
\subfloat[]{
\begin{minipage}[t]{0.33\linewidth}
\centering
\includegraphics[width=2.2in]{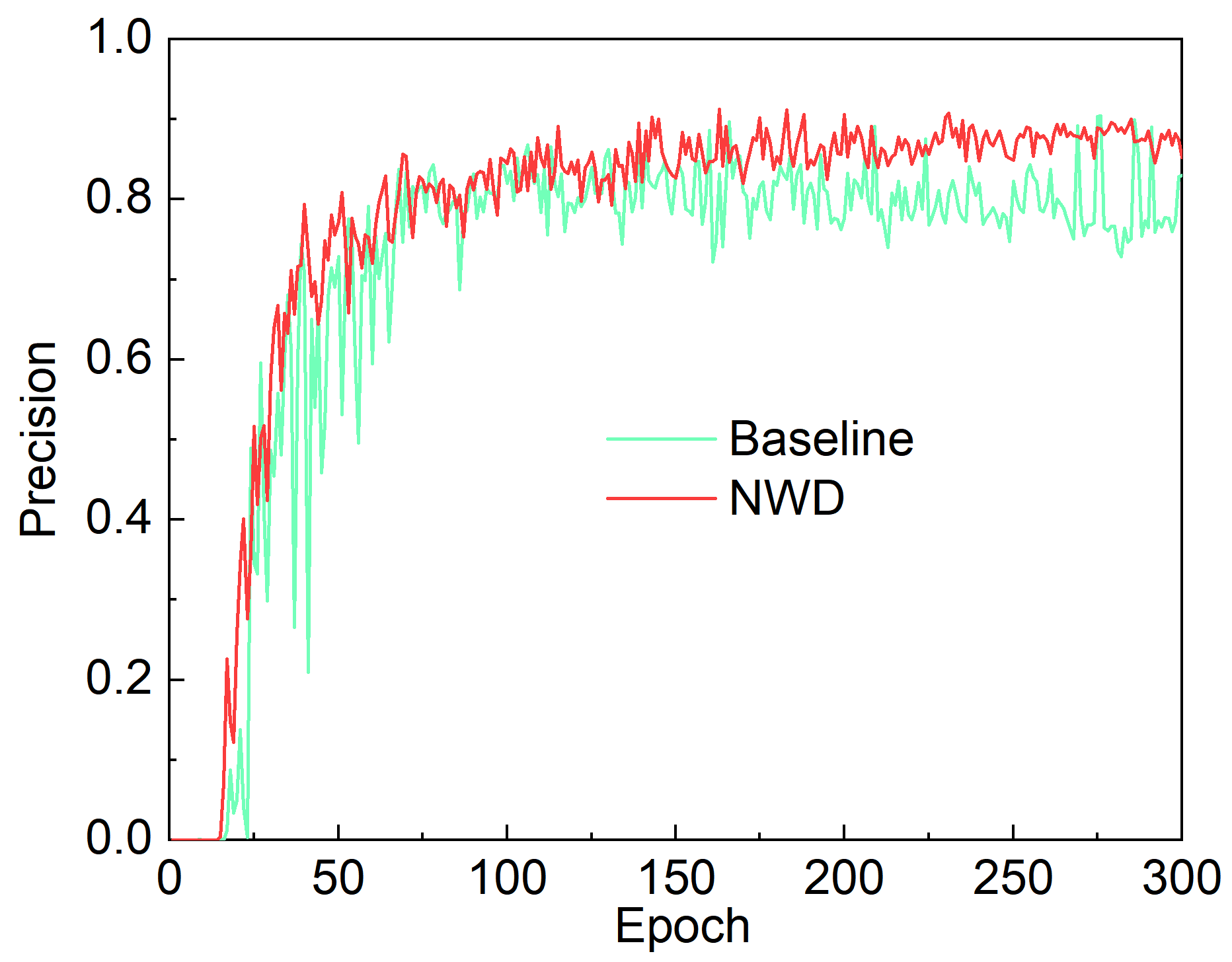}
\end{minipage}%
}%
\subfloat[]{
\begin{minipage}[t]{0.33\linewidth}
\centering
\includegraphics[width=2.2in]{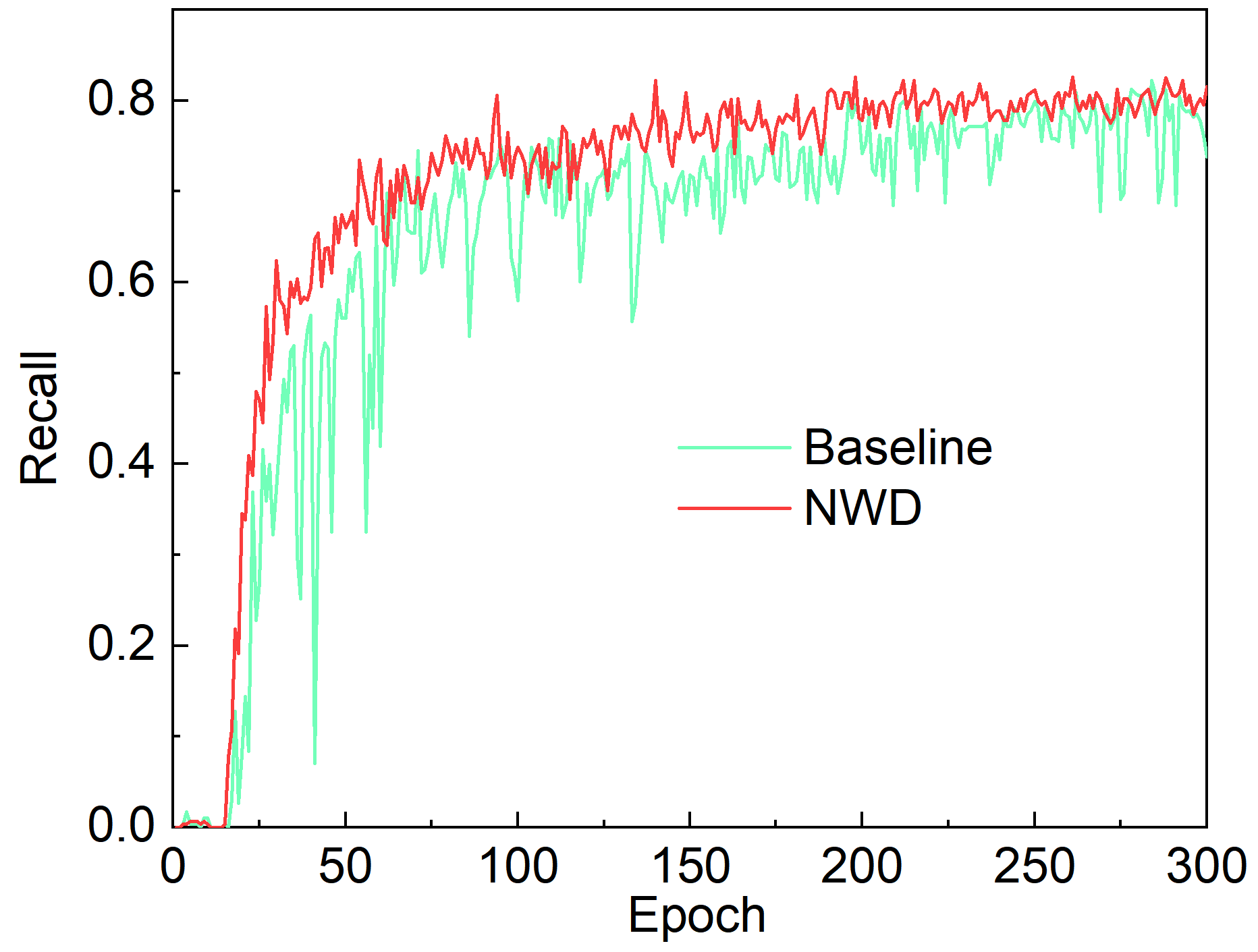}
\end{minipage}%
}%
\subfloat[]{
\begin{minipage}[t]{0.33\linewidth}
\centering
\includegraphics[width=2.2in]{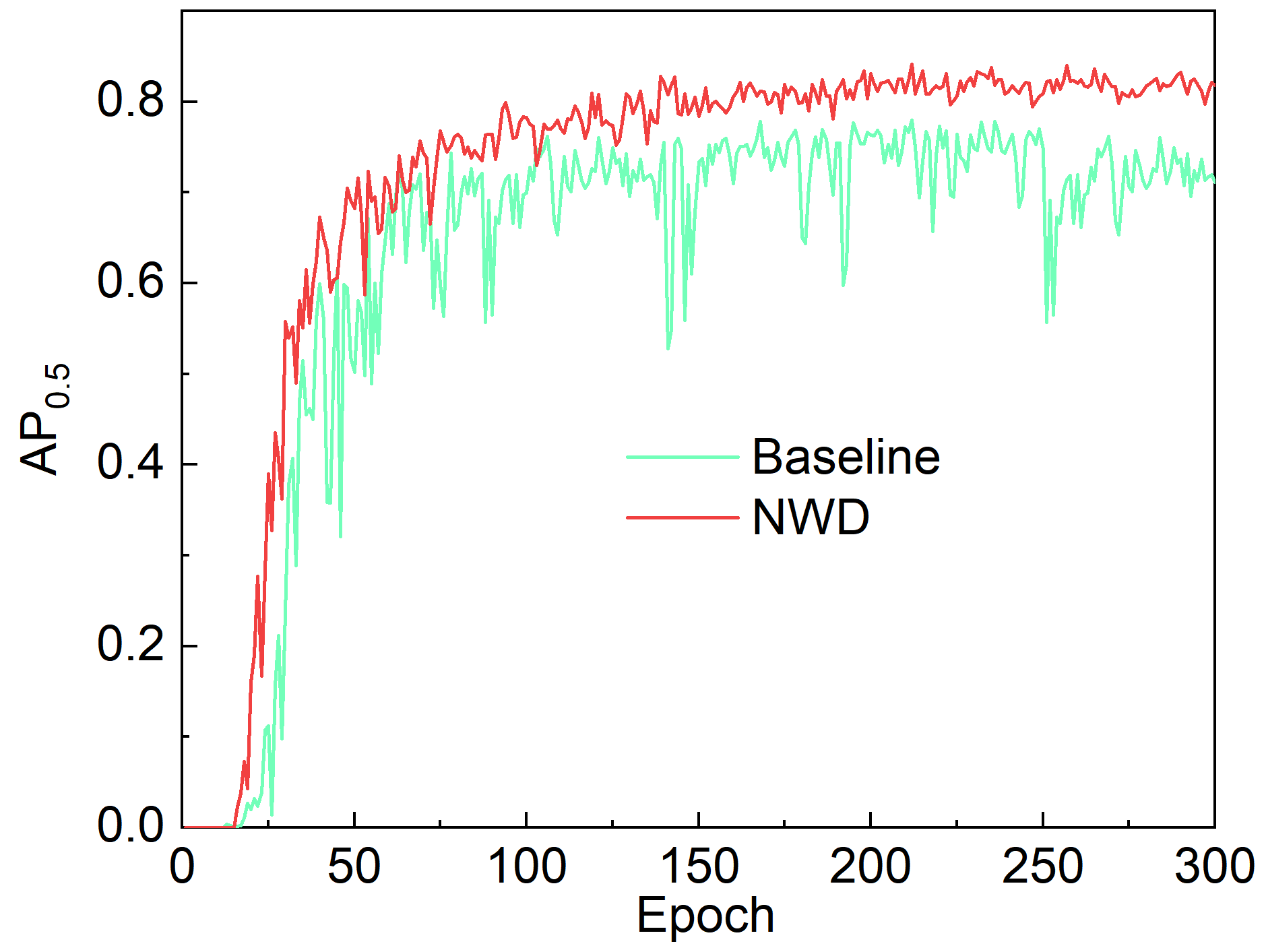}
\end{minipage}
}%
\centering
\caption{The baseline and NWD comparison in terms of precision, recall and $AP_{0.5}$ in training process. (a) Precision comparison. (b) Recall comparison. (c) $AP_{0.5}$ comparison.}
\label{fig_7}
\end{figure*}

\textbf{Impact of NWD:} As mentioned previously, the parameter $C$ is closely tied to the dataset. In order to investigate its influence on the model, we conducted experiments by varying the value of parameter $C$ as shown in Table \ref{table3}. The NWD enhances the quality of both positive and negative samples during the training process. As a result, the application of NWD yields a significant improvement in the model's performance, and reaching its optimum at $C=11$. Fig. \ref{fig_7} illustrates the changes in evaluation metrics during the training process. Since the IoU metric is particularly sensitive to infrared small target, leading to similarities between positive and negative samples. Accordingly, the model encounters difficulties in convergence, resulting in substantial fluctuations in the evaluation metric. However, as can be seen from the figure, the integration of NWD alleviates the problem of difficult in model convergence.
\begin{table}[H]
\centering
\caption{ Ablation study on the different parameter $C$ of NWD in precision, recall, $AP_{0.5}$ and F1.\label{tab:table3}}
\begin{tabular}{|c|c|c|c|c|}
\hline
$C$        & $Precision$   & $Recall$   & $AP_{0.5}$   & $F1$    \\ \hline
Baseline & 0.845 & 0.756 & 0.770 & 0.798 \\ \hline
9        & 0.867 & 0.775 & 0.789 & 0.818 \\ \hline
11       & 0.890 & 0.781 & 0.806 & 0.832 \\ \hline
13       & 0.867 & 0.797 & 0.800 & 0.831 \\ \hline
15       & 0.864 & 0.778 & 0.795 & 0.819 \\ \hline
17       & 0.899 & 0.771 & 0.801 & 0.830 \\ \hline
\end{tabular}
\label{table3}
\end{table}

\textbf{Impact of dynamic head:} It can be seen that IoU metric can easily lead to the model not convergence, and it is difficult to evaluate the actual effect of the network. Therefore, we conducted experiments on dynamic head under the premise of using NWD. Table \ref{table4} shows the obvious improvement in model performance after incorporating the dynamic head module. Moreover, as the quantity of dynamic head is augmented, the performance of the model will increase slightly. The optimal performance result can be achieved when the number of dynamic head module is 4.

\begin{table}[H]
\centering
\caption{ Ablation study on the different number of dynamic head blocks in precision, recall, $AP_{0.5}$ and F1.\label{tab:table4}}
\begin{tabular}{|c|c|c|c|c|}
\hline
Block & $Precision$ & $Recall$ & $AP_{0.5}$ & $F1$    \\ \hline
0     & 0.867     & 0.775  & 0.789 & 0.818 \\ \hline
1     & 0.850     & 0.804  & 0.797 & 0.826 \\ \hline
2     & 0.896     & 0.772  & 0.794 & 0.829 \\ \hline
3     & 0.881     & 0.788  & 0.792 & 0.832 \\ \hline
4     & 0.861     & 0.814  & 0.799 & 0.837 \\ \hline
5     & 0.860     & 0.814  & 0.800 & 0.836 \\ \hline
\end{tabular}
\label{table4}
\end{table}

In addition to analyzing the effectiveness of each component, we also experimented with the combined effects of multiple components. It can be seen from the Table \ref{table5} that only the dyhead raises the number of network parameters, and the design of the loss function does not additional increase the complexity of the model. Moreover, the performance can be improved with the addition of each component, indicating that the designed loss function and the network structure can be well integrated.
\begin{table*}[!htbp]
\centering
\caption{ Ablation study on the ATFL, NWD and dynamic head in precision, recall, $AP_{0.5}$, F1 and GFLOPs.\label{tab:table5}}
\begin{tabular}{|c|c|c|c|c|c|c|c|c|}
\hline
Dataset                & ATFL & NWD & Dynamic head & $Precision$ & $Recall$ & $AP_{0.5}$  & F1    & Parameters(M) \\ \hline
\multirow{4}{*}{IRSTD} & $\times$    & $\times$   & $\times$      & 0.845     & 0.756  & 0.770 & 0.798 & 32.769 \\   \cline{2-9} 
                       & $\checkmark$    & $\times$   & $\times$      & 0.851     & 0.768  & 0.773 & 0.807 & 32.769 \\ \cline{2-9}
                       & $\checkmark$    & $\checkmark$   & $\times$      & 0.858     & 0.797  & 0.801 & 0.826 & 32.769 \\ \cline{2-9}
                       & $\checkmark$    & $\checkmark$   & $\checkmark$      & \textbf{0.882}     & \textbf{0.807}  & \textbf{0.816} & \textbf{0.843} & 38.519 \\ \hline
\end{tabular}
\label{table5}
\end{table*}

\section{Conclusion}\label{section5}
This paper presented the EFLNet, an innovative approach aimed at enhancing the feature learning capability of infrared small target, thereby improving the performance of infrared small target detection. Specifically, we designed a novel ATFL loss function that automatically adjusted the loss weights, allowing for differentiated treatment of the target and background, which alleviated the inherent imbalance problem between the target and background within the image. The NWD metric facilitated the generation of superior quality positive and negative samples, effectively resolving the sensitivity issues associated with the IoU metric when dealed with infrared small target. By leveraging dynamic head, the relative importance of each semantic layer can be learned, and more attention was paid to the shallow features of infrared small target. Experiments on public datasets showed that our method outperforms SOTA methods. Additionally, we provided the additional bounding box annotation forms of the existing infrared small target datasets, which makes it possible to treat infrared small target detection as a detection-based task.

\bibliographystyle{IEEEtran}
\bibliography{reference.bib}
\begin{IEEEbiography}[{\includegraphics[width=1in,height=1.25in,clip,keepaspectratio]{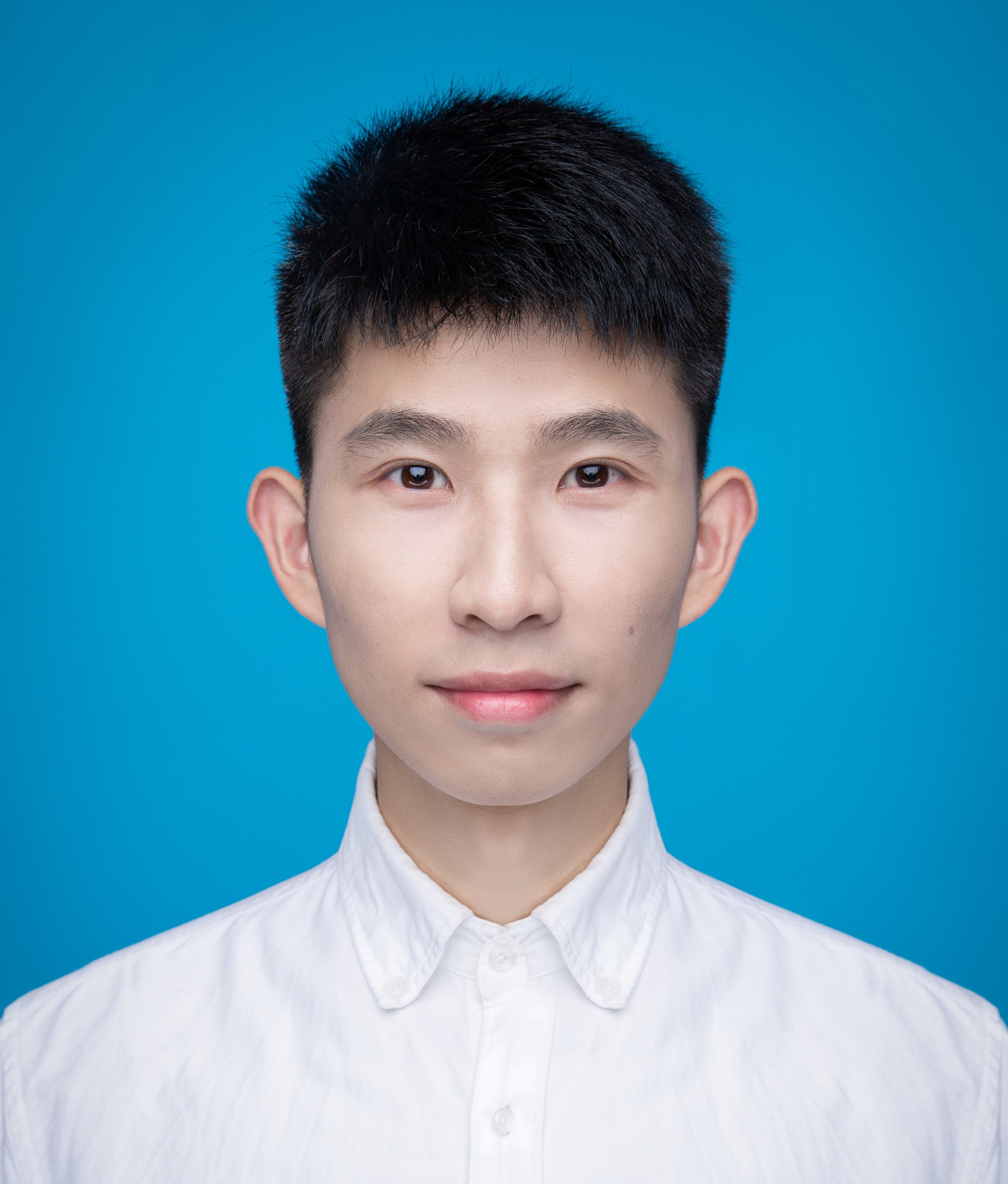}}]{Bo Yang}
was born in 1995. He received the M.S. degree from the Chongqing Jiaotong University, Chongqing, China, in 2018. He is currently pursuing the Ph.D. degree with College of Mechanical and Vehicle Engineering, State Key Laboratory of Mechanical Transmission, Chongqing University. His research interests include deep learning, target detection,intelligent unmanned system,and target tracking.
\end{IEEEbiography}
\begin{IEEEbiography}[{\includegraphics[width=1in,height=1.25in,clip,keepaspectratio]{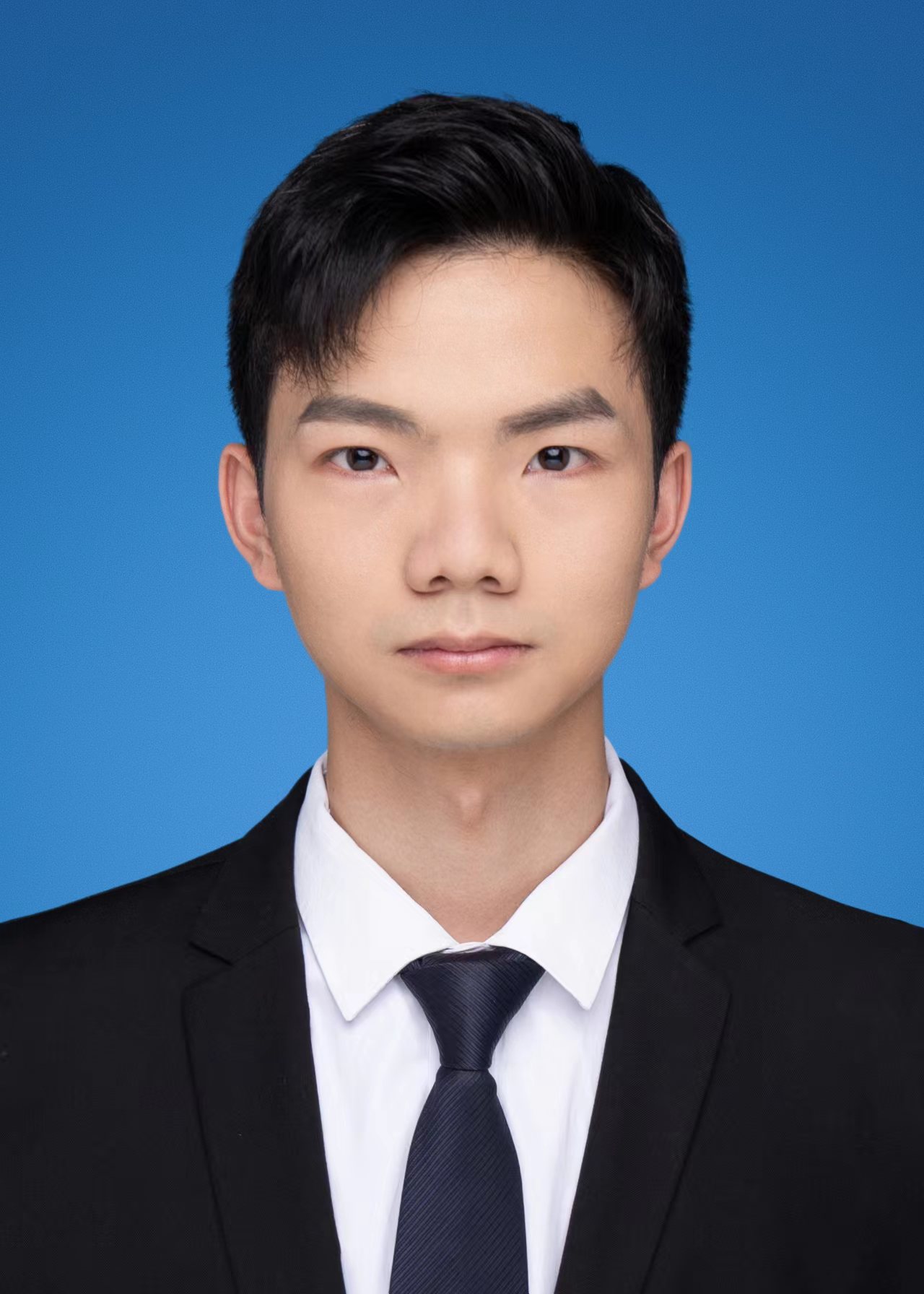}}]{Xinyu Zhang}
Xinyu Zhang received his B.S. degree from School of Mechanical Engineering, Tiangong University of China in 2021. Currently, he is pursuing the M.S. degree in School of Mechanical and Vehicle Engineering, Chongqing University of China. His research interests include Maneuver trajectory prediction,,intelligent unmanned system, and Intention recognition.
\end{IEEEbiography}
\begin{IEEEbiography}[{\includegraphics[width=1in,height=1.25in,clip,keepaspectratio]{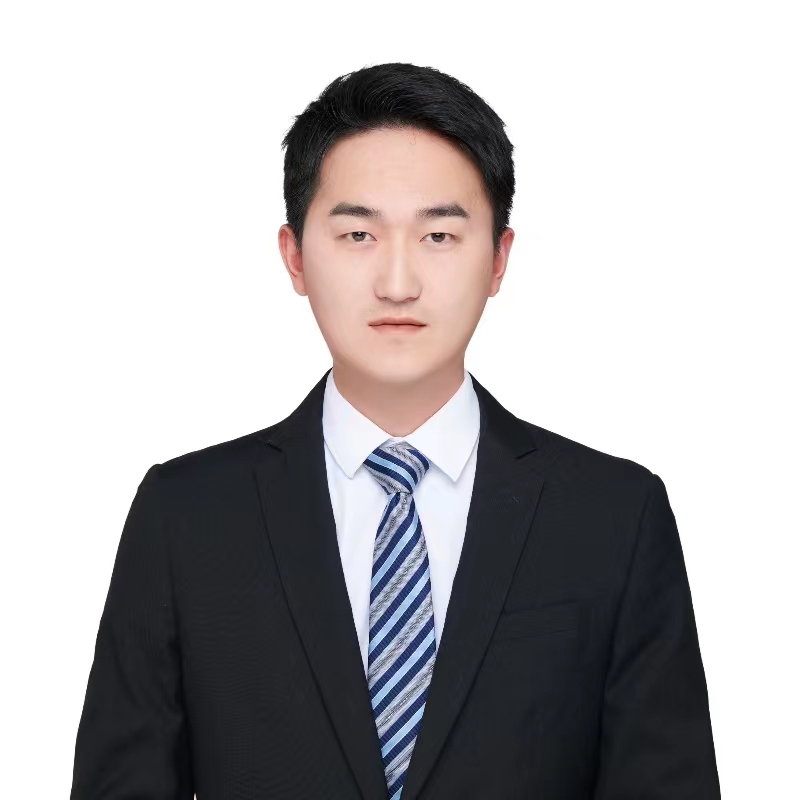}}]{Jian Zhang}
received the B.S. degree in School of Civil Engineering from Chongqing Jiaotong University, Chongqing, China, in 2014, and the M.S. degree in School of Aerospace Engineering and Applied Mechanics from Tongji University, Shanghai, China, in 2017. From 2017 to 2018, he was an Assistant engineer of State Key Laboratory of Vehicle NVH and Safety Technology, Chongqing, China. He is currently pursuing the Ph.D. degree in the College of Mechanical and Vehicle Engineering, Chongqing University, Chongqing, China. His research interests include nonlinear dynamics, nonlinear control, distributed parameter system, and flexible robotics.
\end{IEEEbiography}
\begin{IEEEbiography}[{\includegraphics[width=1in,height=1.25in,clip,keepaspectratio]{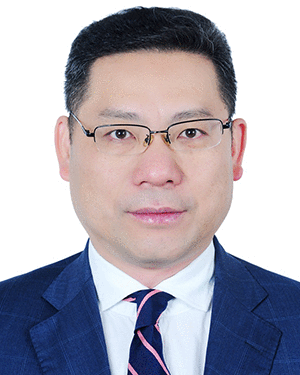}}]{Jun Luo}
received the B.S. and M.S. degrees in mechanical engineering from Henan Polytechnic University, Jiaozuo, China, in 1994 and 1997, respectively, and the Ph.D. degree in mechanical engineering from the Research Institute of Robotics, Shanghai Jiao Tong University, Shanghai, China, in 2000.,He is currently a Professor with the State Key Laboratory of Mechanical Transmissions, Chongqing University, Chongqing, China. His current research interests include artificial intelligence, sensing technology,,intelligent unmanned system, and special robotics.
\end{IEEEbiography}
\begin{IEEEbiography}[{\includegraphics[width=1in,height=1.25in,clip,keepaspectratio]{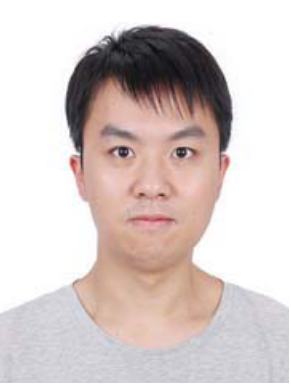}}]{Mingliang Zhou}
(Member, IEEE) received the Ph.D. degree in computer science from Beihang University, Beijing, China, in 2017. He was a Postdoctoral Researcher with the Department of Computer Science, City University of Hong Kong, Hong Kong, from September 2017 to September 2019. He is currently a Lecturer with the School of Computer Science, Chongqing University, Chongqing, China. He is also a Postdoctoral Researcher with the State Key Laboratory of Internet of Things for Smart City, University of Macau. His research interests include image and video coding, perceptual image processing, multimedia signal processing, rate control, multimedia communication, machine learning, and optimization.
\end{IEEEbiography}
\begin{IEEEbiography}[{\includegraphics[width=1in,height=1.25in,clip,keepaspectratio]{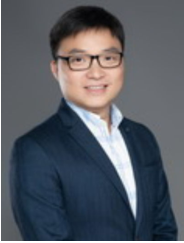}}]{Yangjun Pi}
received the B.Eng.degree in Mechatronic Engineering and the Ph.D. degree in Mechanical Engineering from Zhejiang University, Hangzhou, China, in 2005 and 2010 respectively.He is currently a Professor in the the State Key Laboratory of Mechanical Transmissions and the College of Mechanical and Vehicle Engineering, Chongqing University, Chongqing, China. His research interests include control of distributed parameter systems,,intelligent unmanned system, and vibration control.
\end{IEEEbiography}
\end{document}